\documentclass[letterpaper]{article} 
\usepackage[preprint]{aaai2027} 
\usepackage[hyphens]{url} 
\usepackage{graphicx} 
\usepackage{natbib} 
\usepackage{caption} 
\usepackage{amsmath}
\usepackage{amssymb}
\usepackage{booktabs}
\usepackage{colortbl}
\usepackage{tabularx}
\usepackage{array}
\usepackage{arydshln}
\usepackage{tikz}
\usepackage{dblfloatfix}

\newcolumntype{L}{>{\raggedright\arraybackslash}X}
\newcolumntype{Y}{>{\centering\arraybackslash}X}

\definecolor{rowgraystrong}{gray}{0.78}
\definecolor{rowgraydark}{gray}{0.84}
\definecolor{rowgraymid}{gray}{0.89}
\definecolor{rowgraylight}{gray}{0.94}
\definecolor{rowgrayfaint}{gray}{0.98}

\renewcommand{\topfraction}{0.95}
\renewcommand{\dbltopfraction}{0.95}
\renewcommand{\bottomfraction}{0.70}
\renewcommand{\textfraction}{0.05}
\title{CoWAM: Coordination Contracts for Selective Policy Intervention with WAMs}
\makeatletter
\def\showauthors@on{T}
\def\copyright@on{}
\let\cowam@aaai@maketitle\@maketitle
\def\@maketitle{\cowam@aaai@maketitle\vspace{-0.12in}}
\makeatother
\author{\normalfont\normalsize
Shuaijun Liu\textsuperscript{1}, Qifu Wen\textsuperscript{2,3},
Shuyang Hao\textsuperscript{1}, Qi Luo\textsuperscript{1}\\
Chenglong Zhang\textsuperscript{1}, Feiyang You\textsuperscript{1},
Chengyu Wu\textsuperscript{1}, Ningxin Su\textsuperscript{1,*}}
\affiliations{\normalfont\small
\textsuperscript{1}The Hong Kong University of Science and Technology (Guangzhou)\\
\textsuperscript{2}Boston University \quad
\textsuperscript{3}Shanghai Jiao Tong University\\
\textsuperscript{*}Corresponding author}

\begin{document}
\maketitle

\begin{abstract}
World Action Models (WAMs) augment robot policies with action-conditioned
predicted futures, but a plausible future alone does not justify changing the
action that a bimanual policy would execute.
We present CoWAM, a selective intervention layer that expresses
synchronization, role compatibility, and collision convergence as
\emph{coordination contracts}.
Each contract combines typed admissibility checks with event-conditioned
verification and calibrated intervention gates.
CoWAM preserves the nominal action unless an alternative satisfies every
active obligation and provides a clear, low-risk improvement; when the nominal
action is also inadmissible, it invokes a predefined abstention fallback.
To separate selector quality from proposal quality, all methods operate on
identical candidate pools and commit their decisions before shared oracle
labeling.
Across eight simulated bimanual tasks, CoWAM improves coordination-valid
selection by 16.7 percentage points over the contract-only variant and raises
closed-loop success by 9.6 percentage points over the strongest selective
baseline, while keeping harmful interventions below 1\%.
Together, these results establish coordination contracts as an effective
interface for conservative policy intervention with predicted world-action
evidence across coordination-rich bimanual tasks.
\end{abstract}

\section{Introduction}

\begin{figure}[t]
\centering
\includegraphics[width=\linewidth]{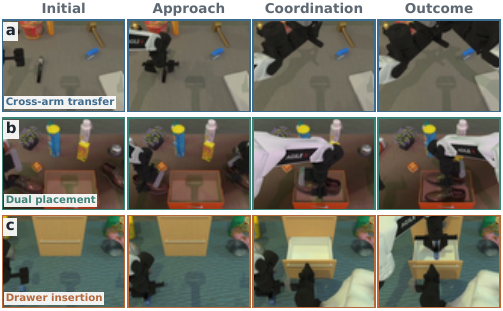}
\caption{\textbf{Coordination-rich task structures.}
RoboTwin~2.0 examples span cross-arm transfer, parallel placement, and
multi-stage insertion beyond simple pick-and-place, exposing synchronization,
role-assignment, spatial-compatibility, and phase-consistency obligations.}
\label{fig:intro-task-scope}
\end{figure}

Bimanual manipulation couples two action streams through shared objects,
timing, workspace, and arm roles.
Action-chunking policies produce coherent paired motion
\citep{Zhao-RSS-23,Chi-RSS-23}, while world action models (WAMs) additionally
pair proposed actions with predicted consequences
\citep{zhu2025uwm,li2025uva,yuan2026fastwam,guo2026xwam}.
These futures can expose failures before execution, but visual plausibility
alone does not determine when an alternative should replace
\emph{Policy Top-1} ($i=0$), the proposer's first-ranked candidate and
hereafter the nominal action.
A coherent rollout may still contain a delayed grasp, incompatible arm
assignment, or converging collision; an unsupported override can likewise
turn uncertainty into failure.
Our setting therefore extends beyond isolated pick-and-place:
Figure~\ref{fig:intro-task-scope} highlights cross-arm transfer, parallel
object placement, and multi-stage insertion, which expose
coordination-validity questions in synchronization, role assignment, spatial
compatibility, and phase consistency.

\begin{figure*}[t]
\centering
\includegraphics[width=\textwidth]{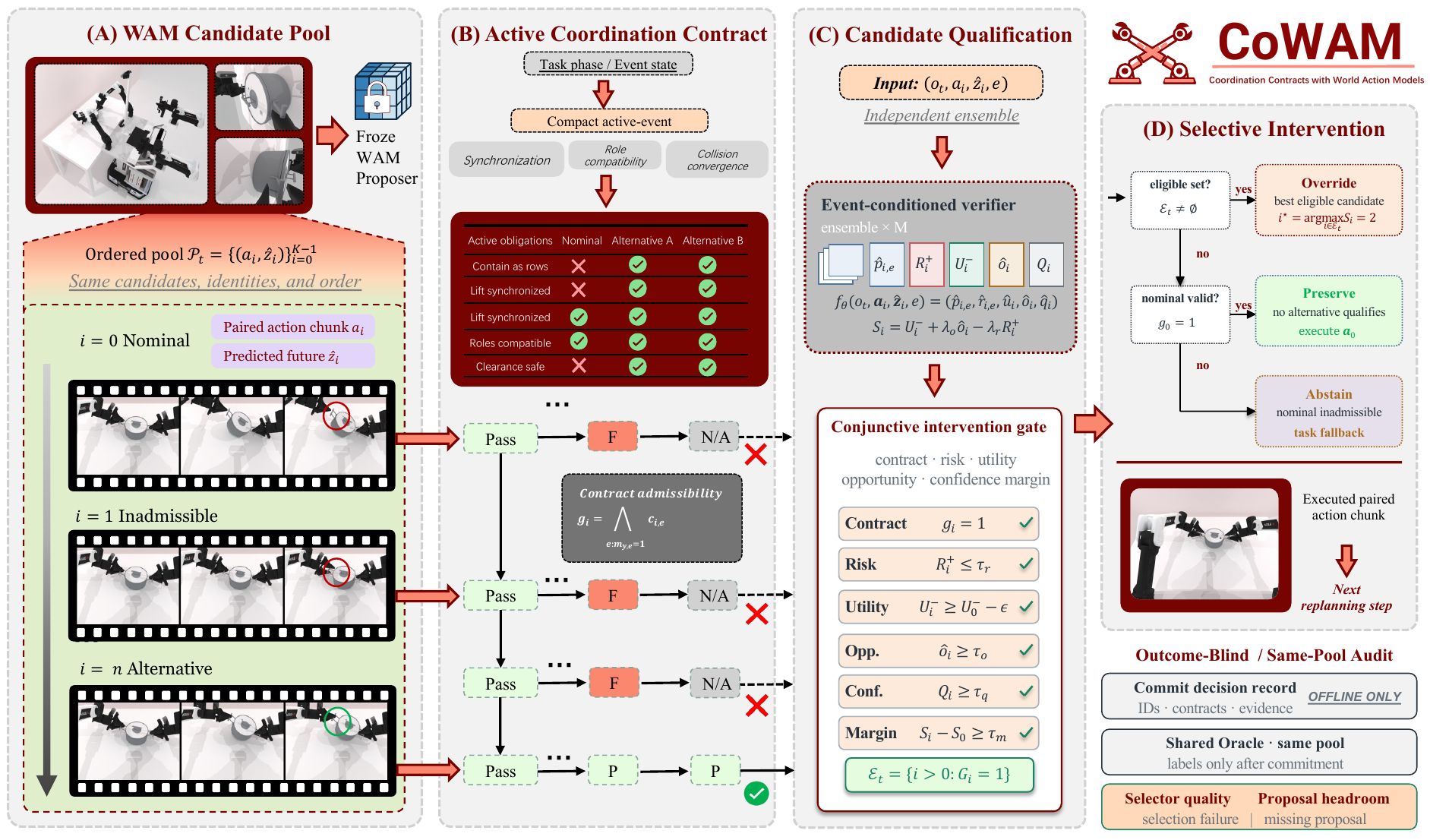}
\caption{\textbf{CoWAM overview and method framework.}
A frozen WAM supplies action-future candidates; coordination contracts and
calibrated evidence support selective preserve, override, or abstain decisions.}
\label{fig:method-overview}
\end{figure*}

We introduce \textbf{CoWAM}, a selective intervention layer that represents
synchronization, role, and collision obligations as \emph{coordination
contracts}.
Each contract combines typed admissibility predicates, event-conditioned
learned evidence, and calibrated intervention gates.
The nominal action remains in effect unless an alternative satisfies every
active obligation, remains low-risk, preserves task utility, and clears the
calibrated intervention thresholds.
Otherwise CoWAM preserves the nominal action or invokes a predefined
abstention fallback.

This contract view separates proposal generation from intervention.
CoWAM neither trains a new action generator nor changes the candidate pool.
Every selector instead receives the same ordered candidates, commits before
simulator outcomes are available, and is audited by one shared oracle-labeling
pass.
We refer to this protocol as \emph{outcome-blind same-pool evaluation}.
The evaluation therefore distinguishes coordination-valid selection, false and
harmful intervention, natural closed-loop success, and proposal headroom.

Across eight RoboTwin tasks and three coordination event families, CoWAM
converts 140 of 150 opportunities versus 115 for the contract-only variant,
with five false and one harmful intervention among 180 matched negatives.
It also records 151 successes across 240 natural episodes, compared with 128
for the strongest selective baseline, with a positive gain on every evaluated
task.
Our contributions are: (1) \textbf{coordination contracts} that specify the
obligations and evidence required for WAM-based intervention; (2) a
\textbf{contract-conditioned selective controller} with typed checks,
event-specific verification, calibrated bounds, and decisions to preserve,
override, or abstain; (3) an \textbf{outcome-blind same-pool evaluation} that separates
selector quality from proposal headroom; and (4) mechanism and robustness
evidence across component removals, candidate counts, event families,
sequential horizons, and proposer sources.

\section{Related Work}

Robot WAMs increasingly couple predicted observations or latent states with
action generation.
Unified World Models and UVA jointly model video and actions
\citep{zhu2025uwm,li2025uva}; Fast-WAM studies the role of test-time
imagination \citep{yuan2026fastwam}; X-WAM predicts multi-view RGB-D futures
\citep{guo2026xwam}; and DreamZero executes a WAM as a closed-loop policy
\citep{ye2026dreamzero}.
CoWAM addresses the complementary question of whether an existing WAM
candidate provides sufficient evidence to replace the nominal action.

Predicted futures also support policy steering and verification.
FOREWARN aligns action-conditioned latent futures with a vision-language model
\citep{WuY1-RSS-25}; future-compatibility scoring tests action-outcome
agreement \citep{ruan2026futurecompatible}; and adaptive execution compares
observations with imagined rollouts \citep{wang2026trustimagination}.
Selective prediction and calibrated ensembles provide related tools for
abstention and uncertainty
\citep{geifman2017selective,guo2017calibration,lakshminarayanan2017ensembles}.
CoWAM instead qualifies candidate interventions through typed event contracts
and calibrated conjunctive gates.

Bimanual policies must preserve timing, object roles, and collision
constraints.
ACT, Diffusion Policy, and 3D Diffusion Policy model paired or multimodal
actions \citep{Zhao-RSS-23,Chi-RSS-23,Ze-RSS-24}; ALOHA Unleashed and RDT-1B
scale contact-rich bimanual learning \citep{zhao2025aloha,liu2025rdt}; and
RoboTwin supplies dual-arm simulation tasks
\citep{Mu_2025_CVPR,chen2025robotwin}.
Unlike constraint-integrated action generation \citep{BouvierJ-RSS-25}, CoWAM
keeps both policy and WAM frozen and validates candidate coordination before
intervention.

\section{Methods}

\subsection{World-Action Candidate Interface}

At replanning time $t$, a frozen proposer returns an ordered candidate pool
\begin{equation}
 \mathcal P_t=\{(\mathbf a_i,\hat{\mathbf z}_i)\}_{i=0}^{K-1},
 \label{eq:pool}
\end{equation}
where $\mathbf a_i$ is a synchronized left-right action chunk,
$\hat{\mathbf z}_i$ is its predicted world trajectory, and $i=0$ denotes
the nominal candidate.
The prediction may include multi-view RGB-D observations and proprioception.
CoWAM neither resamples this pool nor changes the proposer.
Candidate identities and order remain fixed through verification, selection,
execution, and audit.
The interface consumes synchronized action candidates and
candidate-conditioned evidence without coupling the selector to the
proposer's generation objective.
In our realization, multi-view RGB-D predictions and proprioception instantiate
the contract fields directly.
The same contract interface and calibrated selector operate across all
proposer sources evaluated in Experiments.

The controller returns one of three decisions.
\emph{Preserve} executes the nominal action, \emph{override} executes one verified
alternative, and \emph{abstain} invokes a task-defined fallback when no
candidate is admissible.
This makes policy intervention, rather than future generation, the object of
the method.

\subsection{Coordination Contracts}

A coordination contract specifies which event obligations are active and what
evidence is required for intervention.
We use three typed families: synchronization, role compatibility, and
collision convergence.
Let $\mathcal E$ be the event vocabulary, $m_{t,e}\in\{0,1\}$ indicate whether
event $e$ is active, and $c_{i,e}\in\{0,1\}$ denote the corresponding
deterministic predicate for candidate $i$.
The contract-admissibility indicator is
\begin{equation}
 g_i=\prod_{e\in\mathcal E}
 \left(1-m_{t,e}+m_{t,e}c_{i,e}\right).
 \label{eq:contract}
\end{equation}
Thus inactive event types impose no constraint, whereas every active
predicate must pass.
Predicates use paired action and future evidence: examples include bounded
inter-arm distance, compatible object assignments, synchronized contact
progress, and nondivergent shared-object motion.
The contract also stores calibrated decision thresholds and the fallback
associated with contract failure.

Contracts are deliberately typed rather than collapsed into one plausibility
score.
They expose why a candidate is inadmissible, determine which predictions are
relevant to the current task phase, and preserve a valid nominal action unless
an alternative supplies positive evidence for intervention.
They also impose three invariants.
First, contract activation is determined before candidate outcomes are known.
Second, every active obligation is evaluated for every candidate under the
same information boundary.
Third, failure of one required obligation cannot be compensated by an
unrelated high utility score.
These invariants prevent a productive-looking candidate from hiding a
specific coordination violation.
Throughout this paper, a coordination contract denotes this complete decision
object rather than predicates in isolation: typed obligations define
admissibility, event-conditioned evidence estimates future satisfaction, and
calibrated gates determine whether that evidence is strong enough to replace
the nominal action. The Contract-only ablation retains the typed predicates
while removing the learned evidence and gate stack.

\subsection{Event-Conditioned Contract Evidence}

Deterministic predicates capture necessary structure but cannot resolve every
future-dependent failure.
An event-conditioned verifier therefore maps the current observation,
candidate action, predicted future, and active event to
\begin{equation}
 f_\theta(o_t,\mathbf a_i,\hat{\mathbf z}_i,e)
  =(\hat p_{i,e},\hat r_{i,e},\hat u_i,\hat o_i,\hat q_i),
 \label{eq:verifier}
\end{equation}
where $\hat p_{i,e}$ estimates event satisfaction, $\hat r_{i,e}$ estimates
coordination risk, $\hat u_i$ is task utility, $\hat o_i$ is opportunity
value, and $\hat q_i$ is confidence.
The verifier is trained and calibrated on task-seed-disjoint groups.
Independently seeded models provide predictive variation.
For aggregate risk and utility, CoWAM constructs conservative bounds
\begin{align}
 R_i^{+}&=\bar r_i+\kappa_r s_i^r,&
 U_i^{-}&=\bar u_i-\kappa_u s_i^u,
 \label{eq:bounds}
\end{align}
where bars denote ensemble means, $s_i^r$ and $s_i^u$ denote predictive
dispersion, and $\kappa_r,\kappa_u$ are frozen calibration multipliers.
We define $Q_i$ as the minimum satisfaction confidence over active events and
use the frozen score
\begin{equation}
 S_i=U_i^{-}+\lambda_o\hat o_i-\lambda_r R_i^{+},
 \label{eq:score}
\end{equation}
with nonnegative coefficients fixed before evaluation.
All normalization, ensemble members, calibration multipliers, and thresholds
are fixed on task-seed-disjoint training and validation groups.
Test groups are used once for the reported discrimination and calibration
metrics.
Conditioning on $e$ allows the same predicted motion to be interpreted
differently when the relevant obligation is synchronization, role assignment,
or collision convergence.
Scalar verifier removes this distinction while retaining a learned candidate
score.

\subsection{Selective Policy Intervention}

For an alternative $i>0$, all intervention conditions are combined as
\begin{align}
 G_i={}&g_i\,
 \mathbf 1[R_i^{+}\leq\tau_r]\,
 \mathbf 1[U_i^{-}\geq U_0^{-}-\epsilon_u]\nonumber\\
 &\times\mathbf 1[\hat o_i\geq\tau_o]\,
 \mathbf 1[Q_i\geq\tau_q]\,
 \mathbf 1[S_i-S_0\geq\tau_m].
 \label{eq:gate}
\end{align}
The terms respectively enforce contract validity, bounded risk, utility
retention, an active opportunity, event confidence, and a selective margin
over the nominal action.
If at least one alternative satisfies $G_i=1$, CoWAM chooses the highest-score
candidate, breaking ties by the original proposer order.
If none passes, it preserves the nominal action when $g_0=1$ and abstains through
the contract fallback otherwise.
Opportunity and margin serve different purposes.
The absolute opportunity gate rejects pools in which no alternative is
predicted to be useful, whereas the relative margin rejects changes that are
not decisively better than the nominal candidate.
Utility retention prevents a locally safer motion from discarding task
progress; risk and confidence gates protect against uncertain event
satisfaction.
Because the rule is conjunctive, each accepted override has a complete,
inspectable reason record.

Every decision record contains candidate IDs, contract outcomes, verifier
outputs, uncertainty, and the selected mode.
The record is persisted before any outcome label is available.
A shared simulator pass subsequently labels every candidate for task success,
coordination validity, progress, and failure mode.
An override is \emph{beneficial} when it repairs a nominal failure without
losing another required outcome, \emph{harmful} when it loses a required
outcome, and \emph{false} when no task or coordination outcome supports the
change.
Oracle best-candidate success quantifies the proposal ceiling separately from
online selection.
Because every selector commits first and receives labels from the same outcome
batch, paired comparisons share identical simulator outcomes and success
definitions.

\begin{table*}[!t]
\centering
\scriptsize
\setlength{\tabcolsep}{2.2pt}
\renewcommand{\arraystretch}{1.08}
\definecolor{tagstate}{RGB}{224,228,232}
\definecolor{tagfuture}{RGB}{207,228,243}
\definecolor{taggeometry}{RGB}{207,233,220}
\definecolor{tagcontract}{RGB}{228,216,243}
\definecolor{tagevent}{RGB}{246,222,183}
\definecolor{tagoutcome}{RGB}{241,213,213}
\definecolor{tagrule}{RGB}{226,231,235}
\newcommand{\tabletag}[2]{%
  \tikz[baseline=(tag.base)]{
    \node[
      anchor=base,
      rounded corners=1.6pt,
      fill=#1,
      inner xsep=1.65pt,
      inner ysep=0.35pt,
      outer sep=0pt,
      text height=1.3ex,
      text depth=0.3ex,
      font=\tiny\sffamily
    ] (tag) {#2};
  }}
\setlength{\dashlinedash}{2pt}
\setlength{\dashlinegap}{1.4pt}
\begin{tabularx}{\textwidth}{
  >{\raggedright\arraybackslash}p{0.145\textwidth}
  >{\raggedright\arraybackslash}p{0.16\textwidth}
  >{\raggedright\arraybackslash}p{0.14\textwidth}
  *{6}{Y}
}
\toprule
\rowcolor{rowgraystrong}
\multicolumn{9}{l}{\textbf{Aggregate selector comparison}} \\
\textbf{Selector} & \textbf{Evidence channels} & \textbf{Selection rule} &
\multicolumn{2}{c}{\textbf{Valid selection $\uparrow$}} &
\multicolumn{4}{c}{\textbf{Matched-negative intervention $\downarrow$}} \\
& & & \textbf{$n/N$} & \textbf{Rate} &
\textbf{False $n/N$} & \textbf{False rate} &
\textbf{Harm $n/N$} & \textbf{Harm rate} \\
\midrule
\rowcolor{rowgraymid}
\multicolumn{9}{l}{\emph{Deployable baselines}} \\
Policy Top-1 & \tabletag{tagstate}{order} &
\tabletag{tagrule}{top-1} &
92/150 & 61.3\% & 0/180 & 0.0\% & 0/180 & 0.0\% \\
Future-Consensus & \tabletag{tagfuture}{future} &
\tabletag{tagrule}{agreement rank} &
101/150 & 67.3\% & 112/180 & 62.2\% & 16/180 & 8.9\% \\
Static collision gate &
\tabletag{taggeometry}{geometry}\ \tabletag{tagfuture}{future} &
\tabletag{tagrule}{collision reject} &
96/150 & 64.0\% & 39/180 & 21.7\% & 8/180 & 4.4\% \\
RGB-D selector &
\tabletag{taggeometry}{RGB-D}\ \tabletag{tagfuture}{future} &
\tabletag{tagrule}{learned rank} &
126/150 & 84.0\% & 10/180 & 5.6\% & 2/180 & 1.1\% \\
Selective Control &
\tabletag{tagcontract}{partial contract} &
\tabletag{tagrule}{selective} &
112/150 & 74.7\% & 18/180 & 10.0\% & 2/180 & 1.1\% \\
\hdashline
\rowcolor{rowgraymid}
\multicolumn{9}{l}{\emph{CoWAM components}} \\
Contract-only & \tabletag{tagcontract}{contract} &
\tabletag{tagrule}{contract-rank} &
115/150 & 76.7\% & 20/180 & 11.1\% & 3/180 & 1.7\% \\
Scalar verifier &
\tabletag{tagcontract}{contract}\ \tabletag{tagstate}{scalar} &
\tabletag{tagrule}{scalar-gate} &
104/150 & 69.3\% & 31/180 & 17.2\% & 5/180 & 2.8\% \\
\hdashline
\rowcolor{rowgraymid}
\multicolumn{9}{l}{\emph{Proposed method and offline reference}} \\
\textbf{CoWAM} &
\tabletag{tagcontract}{contract}\ \tabletag{tagevent}{event}\
\tabletag{tagfuture}{future} &
\tabletag{tagrule}{full-gate} &
\textbf{140/150} & \textbf{93.3\%} &
\textbf{5/180} & \textbf{2.8\%} &
\textbf{1/180} & \textbf{0.6\%} \\
Oracle Upper Bound & \tabletag{tagoutcome}{outcome} &
\tabletag{tagrule}{oracle} &
150/150 & 100.0\% & 0/180 & 0.0\% & 0/180 & 0.0\% \\
\hdashline
\rowcolor{rowgraystrong}
\multicolumn{9}{l}{\textbf{Event-family decomposition}} \\
\textbf{Event} & \multicolumn{2}{l}{\textbf{Coordination obligation}} &
\textbf{Opp.} & \textbf{Policy} & \textbf{Contract} &
\textbf{CoWAM} & \textbf{False} & \textbf{Harm} \\
\midrule
Synchronization &
\multicolumn{2}{l}{Contacts and releases remain temporally compatible} &
50 & 31 & 39 & \textbf{47} & 2/60 & 0/60 \\
Role compatibility &
\multicolumn{2}{l}{Arms retain task-consistent object and support roles} &
50 & 30 & 37 & \textbf{46} & 2/60 & 0/60 \\
Collision convergence &
\multicolumn{2}{l}{Inter-arm motion avoids convergence toward unsafe contact} &
50 & 31 & 39 & \textbf{47} & 1/60 & 1/60 \\
\midrule
\textbf{Total} & \multicolumn{2}{l}{\textbf{All active coordination contracts}} &
\textbf{150} & \textbf{92} & \textbf{115} &
\textbf{140} & \textbf{5/180} & \textbf{1/180} \\
\bottomrule
\end{tabularx}
\caption{\textbf{Coordination-valid intervention.}
The upper block joins the selector definitions from the appendix with the
complete event ledger: valid selection uses 150 oracle-confirmed
opportunities, while false and harmful intervention use 180 matched
contract-valid negatives. The lower block aligns each evaluated event family
with its coordination obligation and reports valid selections plus CoWAM's
matched-negative errors. CoWAM versus Contract-only has 27 versus 2 discordant
pairs ($p=1.6{\times}10^{-6}$).}
\label{tab:main-event}
\end{table*}

\section{Experiments}

\subsection{Setup}

We evaluate CoWAM in RoboTwin~2.0 \citep{chen2025robotwin} on eight bimanual
tasks: Lift Pot, Pick Dual Bottles, Stack Two Bowls, Place Can in Basket, Put
Bottles in Dustbin, Stack Three Bowls, Scan Object, and Hang Mug.
The primary interface provides paired bimanual actions, multi-view RGB-D
futures, and predicted proprioception; separate robustness tests use X-WAM,
LeWorldModel~\citep{maes_lelidec2026lewm}, and mixed proposer pools through the
same candidate interface.
Within each paired unit, all methods receive the same restored state,
observations, ordered actions, predicted futures, and execution horizon.

We compare Policy Top-1; Future-Consensus, which ranks predicted-future
agreement; a static collision gate; an RGB-D selector; and an earlier
Selective Control baseline.
These cover no intervention, aggressive future reranking, fixed geometric
filtering, and selective intervention without the complete coordination
contract.
Contract-only removes learned verification and the full gate stack, Scalar
verifier removes event conditioning, and further ablations isolate each
conservative gate.
The oracle selects the best candidate after outcome labeling and provides an
offline proposal ceiling for the shared candidate pool.

The coordination audit comprises 180 independent event-stress clusters across
eight tasks and three event families.
A frozen oracle identifies 150 positive opportunities; one matched
contract-valid negative per cluster supplies 180 units for measuring false and
harmful intervention.
Natural closed loop uses 30 held-out seeds per task and method: 240 paired
episode pools per method and 1,440 episodes across six methods.
Candidate scaling uses 80 independent restored states for each
$K\in\{4,8,16,32\}$; learned ranking uses 1,200 task-seed-disjoint groups and
9,600 candidate records.

Success requires strict simulator completion, coordination validity requires
all active event obligations, and intervention means selecting $i\neq0$;
false and harmful interventions follow the Methods definitions.
Statistical units are paired event clusters or task-seed episodes, never
correlated candidates, with two-sided exact paired tests for both headline
comparisons.
Thresholds are selected on disjoint validation units and frozen before
outcome-bearing evaluation.
Frozen denominators retain every outcome-bearing evaluation unit; the appendix
specifies allocation and denominator reuse.

\begin{figure*}[!t]
\centering
\begin{minipage}[t]{0.32\textwidth}
  \centering
  \includegraphics[width=\linewidth]{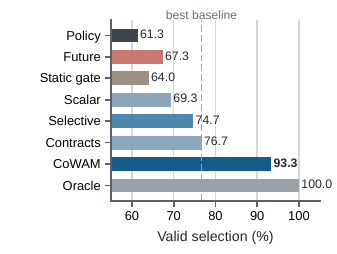}
  \par\vspace{1pt}{\scriptsize\textbf{(a) Valid selection} ($N=150$)}
\end{minipage}
\hfill
\begin{minipage}[t]{0.32\textwidth}
  \centering
  \includegraphics[width=\linewidth]{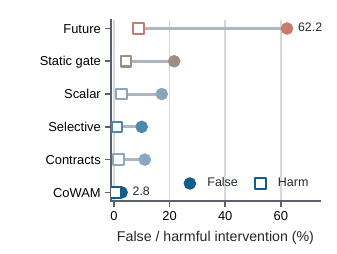}
  \par\vspace{1pt}{\scriptsize\textbf{(b) Active-selector error} ($N=180$)}
\end{minipage}
\hfill
\begin{minipage}[t]{0.32\textwidth}
  \centering
  \includegraphics[width=\linewidth]{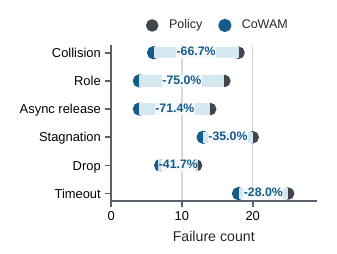}
  \par\vspace{1pt}{\scriptsize\textbf{(c) Failure counts} ($N=240$)}
\end{minipage}
\caption{\textbf{Observed intervention outcomes.}
CoWAM achieves the strongest deployable valid selection, the lowest error
among selectors that intervene, and fewer failures in every recorded category.}
\label{fig:main-outcomes}
\end{figure*}

\begin{figure}[!t]
\centering
\begin{minipage}[t]{0.485\columnwidth}
  \centering
  \includegraphics[width=\linewidth]{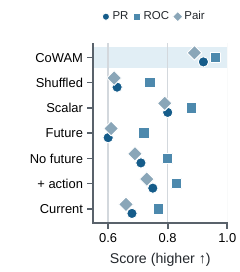}
  \par\vspace{1pt}{\scriptsize\textbf{(a) Ranking discrimination}}
\end{minipage}
\hfill
\begin{minipage}[t]{0.485\columnwidth}
  \centering
  \includegraphics[width=\linewidth]{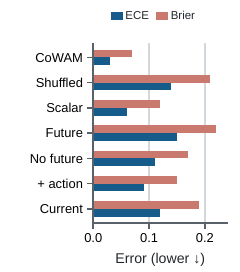}
  \par\vspace{1pt}{\scriptsize\textbf{(b) Calibration error}}
\end{minipage}
\caption{\textbf{Learned contract evidence.}
Event-conditioned CoWAM yields the strongest ranking discrimination and lowest
calibration error among the evaluated representations.}
\label{fig:main-learned-evidence}
\end{figure}

\begin{figure*}[!t]
\centering
\includegraphics[width=\textwidth]{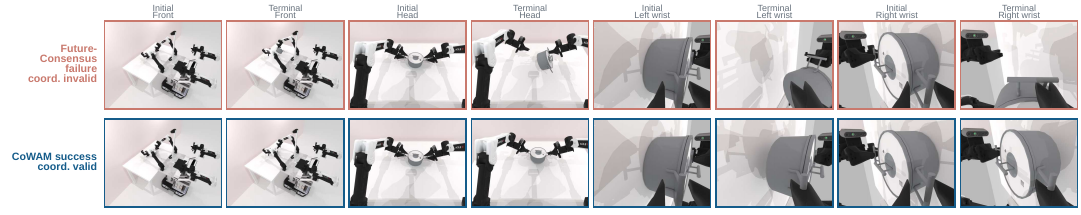}
\caption{\textbf{Lift Pot: CoWAM coordination success.}
From the same restored state and candidate pool, Future-Consensus fails while
CoWAM selects an alternative that achieves task success and coordination
validity across all recorded camera views.}
\label{fig:main-qualitative-lift}
\end{figure*}

\begin{figure}[!t]
\centering
\includegraphics[width=\linewidth]{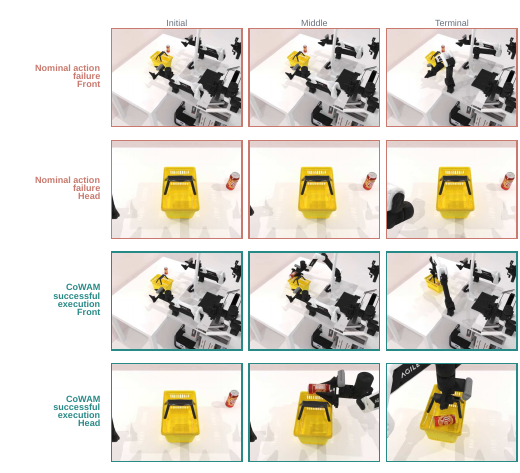}
\caption{\textbf{Place Can in Basket: CoWAM success.}
Initial, intermediate, and terminal views contrast the nominal failure with
CoWAM's successful object acquisition, transport, and basket placement.}
\label{fig:main-qualitative-basket}
\end{figure}

\begin{figure*}[!t]
\centering
\includegraphics[width=\textwidth]{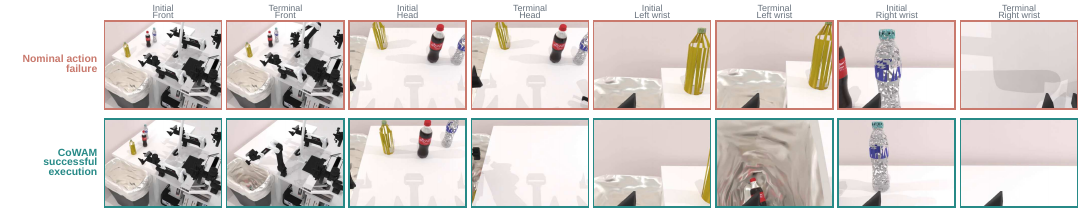}
\caption{\textbf{Put Bottles in Dustbin: CoWAM success.}
Front, head, and wrist views contrast the nominal failure with CoWAM's
successful two-arm object assignment and phase-consistent placement.}
\label{fig:main-qualitative-dustbin}
\end{figure*}

\subsection{Main Results}

CoWAM selects coordination-valid candidates on 140 of 150 opportunities
(93.3\%), compared with 115 of 150 (76.7\%) for Contract-only.
This 16.7 percentage-point gain is supported by 27 versus 2 discordant pairs
($p=1.6{\times}10^{-6}$).
This improvement does not require more frequent or riskier intervention:
CoWAM makes five false and one harmful intervention among 180 matched negatives,
whereas Future-Consensus makes 112 and 16.
The strongest non-CoWAM coordination baseline is the RGB-D selector, with
126 valid selections among 150 opportunities, ten false interventions, and two
harmful interventions among 180 matched negatives.
CoWAM therefore recovers more valid alternatives without increasing
unsupported interventions.
The improvement is consistent across all three event families: CoWAM converts
47 of 50 synchronization, 46 of 50 role-compatibility, and 47 of 50
collision-convergence opportunities, compared with 39, 37, and 39 for
Contract-only.
Higher conversion and lower matched-negative error occur together: event
evidence authorizes rather than merely encourages reranking. The gains in each
family therefore reflect more accurate coordination decisions instead of a
larger intervention budget.

The same method improves natural closed-loop success from 96 of 240 episodes
(40.0\%) for Policy Top-1 and 128 of 240 (53.3\%) for Selective Control to
151 of 240 (62.9\%).
This is a 9.6 percentage-point gain over the strongest selective baseline, with
32 versus 9 discordant pairs ($p=4.3{\times}10^{-4}$).
CoWAM also reaches 95.4\% coordination validity while limiting false and
harmful interventions to 3.3\% and 0.4\%.
Every task contributes a positive gain, demonstrating consistency across the
eight-task evaluation rather than concentration in a single task family.
Together, these gains add 23 successful episodes over Selective Control.
The oracle upper bound succeeds on 176 of 240 pools; CoWAM closes 55 of the
80-success gap between the nominal policy and this proposal ceiling.
Relative to Policy Top-1, CoWAM reduces inter-arm collision from 18 to 6
episodes, role conflict from 16 to 4, and asynchronous release from 14 to 4.
Figure~\ref{fig:main-outcomes} visualizes the resulting validity, intervention
error, and natural failure profile.
Together, the event audit and natural closed loop evaluate complementary
levels of the same claim. The audit tests whether CoWAM identifies
coordination-valid alternatives at intervention opportunities, while natural
episodes test whether those choices translate into complete task execution.

\begin{table}[!t]
\centering
\footnotesize
\setlength{\tabcolsep}{2.2pt}
\renewcommand{\arraystretch}{1.05}
\newcommand{\succcell}[2]{#1\hspace{0.8pt}\raisebox{-0.15ex}{\tiny(#2)}}
\begin{tabularx}{\columnwidth}{
  >{\raggedright\arraybackslash}p{0.28\columnwidth}
  >{\centering\arraybackslash}p{0.22\columnwidth}
  *{3}{Y}}
\toprule
\textbf{Method} & \textbf{Succ.} &
\textbf{Valid $\uparrow$} & \textbf{False $\downarrow$} &
\textbf{Harm $\downarrow$} \\
\midrule
Policy Top-1 & \succcell{96/240}{40.0\%} & 80.8\% & 0.0\% & 0.0\% \\
Future-Consensus & \succcell{111/240}{46.3\%} & 84.6\% & 42.9\% & 5.4\% \\
RGB-D selector & \succcell{120/240}{50.0\%} & 87.9\% & 9.6\% & 2.1\% \\
\rowcolor{rowgraymid}
Selective Control & \succcell{128/240}{53.3\%} & 90.0\% & 5.8\% & 0.8\% \\
\rowcolor{rowgraystrong}
\textbf{CoWAM} & \textbf{\succcell{151/240}{62.9\%}} & \textbf{95.4\%} &
\textbf{3.3\%} & \textbf{0.4\%} \\
Oracle & \succcell{176/240}{73.3\%} & 100.0\% & 0.0\% & 0.0\% \\
\bottomrule
\end{tabularx}
\par\smallskip
\textbf{(a) Aggregate outcomes}
\par\vspace{5pt}
\begin{tabularx}{\columnwidth}{>{\raggedright\arraybackslash}p{0.31\columnwidth}*{4}{Y}}
\toprule
\textbf{Task} & \textbf{Selective} & \cellcolor{white}\textbf{CoWAM} &
\textbf{Success} & \textbf{Oracle} \\
\midrule
Lift Pot & 19/30 & \cellcolor{rowgraylight}\textbf{22/30} & +3 & 24/30 \\
Pick Dual Bottles & 18/30 & \cellcolor{rowgraylight}\textbf{21/30} & +3 & 23/30 \\
Stack Two Bowls & 17/30 & \cellcolor{rowgraylight}\textbf{20/30} & +3 & 23/30 \\
Place Can in Basket & 16/30 & \cellcolor{rowgraylight}\textbf{19/30} & +3 & 22/30 \\
Put Bottles in Dustbin & 14/30 & \cellcolor{rowgraylight}\textbf{18/30} & +4 & 21/30 \\
Stack Three Bowls & 13/30 & \cellcolor{rowgraylight}\textbf{17/30} & +4 & 20/30 \\
Scan Object & 17/30 & \cellcolor{rowgraylight}\textbf{18/30} & +1 & 22/30 \\
Hang Mug & 14/30 & \cellcolor{rowgraylight}\textbf{16/30} & +2 & 21/30 \\
\midrule
\textbf{Total} & \textbf{128/240} &
\cellcolor{rowgraymid}\textbf{151/240} & \textbf{+23} &
\textbf{176/240} \\
\bottomrule
\end{tabularx}
\par\smallskip
\textbf{(b) Per-task consistency}
\caption{\textbf{Natural closed-loop performance.}
(a) Aggregate outcome and intervention quality.
(b) Success consistency across eight tasks, with gain over Selective Control.
Each task uses 30 paired held-out seeds. The aggregate paired discordance is
32 versus 9 ($p=4.3{\times}10^{-4}$).}
\label{tab:main-natural}
\end{table}

\subsection{Ablations}

\begin{table*}[!t]
\centering
\footnotesize
\setlength{\tabcolsep}{3pt}
\renewcommand{\arraystretch}{1.05}
\begin{minipage}[t]{0.49\textwidth}
\vspace{0pt}\centering
\begin{tabularx}{\linewidth}{>{\raggedright\arraybackslash}p{0.49\linewidth}*{3}{Y}}
\toprule
\rowcolor{rowgraydark}
\textbf{Variant} & \textbf{Valid $\uparrow$} &
\textbf{False $\downarrow$} & \textbf{Harm $\downarrow$} \\
\rowcolor{rowgraystrong}
\textbf{Full CoWAM} & \textbf{93.3\%} & \textbf{2.8\%} & \textbf{0.6\%} \\
\rowcolor{rowgraymid}
Contract-only & 76.7\% & 11.1\% & 1.7\% \\
Scalar verifier & 69.3\% & 17.2\% & 2.8\% \\
\rowcolor{rowgraylight}
No event conditioning & 74.7\% & 10.0\% & 1.7\% \\
No depth & 85.3\% & 5.6\% & 1.1\% \\
\rowcolor{rowgraylight}
No uncertainty bound & 94.7\% & 14.4\% & 3.3\% \\
\rowcolor{rowgraymid}
No baseline preservation & 96.7\% & 28.9\% & 7.8\% \\
\bottomrule
\end{tabularx}
\par\smallskip
\textbf{(a) Contract and gate ablation}
\end{minipage}
\hfill
\begin{minipage}[t]{0.49\textwidth}
\vspace{0pt}\centering
\begin{tabularx}{\linewidth}{>{\raggedright\arraybackslash}p{0.38\linewidth}*{3}{Y}}
\toprule
\rowcolor{rowgraydark}
\textbf{Representation} & \textbf{AUPRC $\uparrow$} &
\textbf{Pair acc. $\uparrow$} & \textbf{ECE $\downarrow$} \\
Current only & 0.68 & 0.66 & 0.12 \\
\rowcolor{rowgraylight}
Current + action & 0.75 & 0.73 & 0.09 \\
No future & 0.71 & 0.69 & 0.11 \\
\rowcolor{rowgraylight}
Future-Consensus & 0.60 & 0.61 & 0.15 \\
\rowcolor{rowgraymid}
Scalar verifier & 0.80 & 0.79 & 0.06 \\
\rowcolor{rowgrayfaint}
Future shuffled & 0.63 & 0.62 & 0.14 \\
\rowcolor{rowgraystrong}
\textbf{CoWAM} & \textbf{0.92} & \textbf{0.89} & \textbf{0.03} \\
\bottomrule
\end{tabularx}
\par\smallskip
\textbf{(b) Learned representation ablation}
\end{minipage}
\caption{\textbf{Mechanism and learned evidence.}
(a) Contract and gate variants reuse 150 positive and 180 matched-negative
units. (b) Representation variants use 1,200 task-seed-disjoint groups and
9,600 candidate records. The two panels separate admissibility and
intervention control from learned future-conditioned evidence.}
\label{tab:main-ablation}
\end{table*}

Contract-only trails full CoWAM by 16.7 percentage points, showing that typed
predicates become substantially more effective when combined with
event-conditioned evidence and calibrated intervention gates.
Removing event conditioning reduces validity by 18.7 percentage points, while
the Scalar verifier reaches 69.3\%, confirming the value of obligation-specific
evidence.
Removing uncertainty bounds slightly raises positive selection but increases
false intervention from 2.8\% to 14.4\% and harm from 0.6\% to 3.3\%.
Without baseline preservation, these rates rise to 28.9\% and 7.8\%.
These ablations indicate that strong performance requires both identifying
valid alternatives and controlling when they may replace the nominal action.
The no-depth variant reaches 85.3\% validity, between Contract-only and the
full method. The remaining gain identifies a useful contribution from 3D
correspondence within the method's multimodal evidence.

On task-seed-disjoint groups, event-conditioned CoWAM reaches 0.92 AUPRC,
0.03 expected calibration error, and 0.89 pair accuracy.
Scalar verifier reaches 0.80, 0.06, and 0.79, respectively.
Shuffling predicted futures reduces AUPRC to 0.63, below current-plus-action
features at 0.75.
These comparisons show that the verifier uses candidate-specific temporal
evidence and that event structure improves both discrimination and
calibration.
Figure~\ref{fig:main-learned-evidence} separates ranking discrimination from
calibration error using complementary visual encodings.

At 25\%, 50\%, 75\%, and 100\% selective coverage, CoWAM's observed
coordination-risk violation rates are 0.7\%, 1.2\%, 2.7\%, and 5.8\%.
Future-Consensus rises from 3.3\% to 22.4\%; CoWAM retains the lower violation
rate across the full operating range.

\subsection{Robustness Analysis}

\begin{table}[!t]
\centering
\footnotesize
\setlength{\tabcolsep}{2.2pt}
\renewcommand{\arraystretch}{1.05}
\begin{tabularx}{\columnwidth}{*{5}{Y}}
\toprule
\textbf{$K$} & \textbf{Opp.} & \cellcolor{white}\textbf{CoWAM rescue} &
\textbf{CoWAM false} & \textbf{FC false} \\
\midrule
4 & 28 & \cellcolor{rowgraylight}\textbf{92.9\%} & 5.0\% & 42.5\% \\
8 & 32 & \cellcolor{rowgraylight}\textbf{93.8\%} & 5.0\% & 47.5\% \\
16 & 36 & \cellcolor{rowgraylight}\textbf{94.4\%} & 6.3\% & 52.5\% \\
32 & 40 & \cellcolor{rowgraylight}\textbf{95.0\%} & 6.3\% & 53.8\% \\
\bottomrule
\end{tabularx}
\par\smallskip
\textbf{(a) Candidate-count scaling}
\par\vspace{5pt}
\begin{tabularx}{\columnwidth}{>{\raggedright\arraybackslash}p{0.24\columnwidth}*{5}{Y}}
\toprule
\textbf{Proposer} & {\scriptsize\textbf{Baseline}} &
\cellcolor{white}{\scriptsize\textbf{CoWAM}} & \textbf{Gain} &
\textbf{False} & \textbf{Harm} \\
\midrule
X-WAM & 66/120 & \cellcolor{rowgraylight}\textbf{78/120} &
\cellcolor{rowgrayfaint}\textbf{+10.0\%} & 5/120 & 1/120 \\
LeWorldModel & 60/120 & \cellcolor{rowgraylight}\textbf{75/120} &
\cellcolor{rowgrayfaint}\textbf{+12.5\%} & 7/120 & 1/120 \\
Mixed pool & 70/120 & \cellcolor{rowgraylight}\textbf{85/120} &
\cellcolor{rowgrayfaint}\textbf{+12.5\%} & 6/120 & 1/120 \\
\bottomrule
\end{tabularx}
\par\smallskip
\textbf{(b) Cross-proposer transfer}
\caption{\textbf{Proposal robustness.}
(a) Each candidate count uses 80 independent states; rescue is normalized by
oracle-confirmed opportunities and false intervention by all states.
(b) Each proposer regime uses 120 paired pools. Baseline is the strongest
corresponding non-CoWAM selector; FC denotes Future-Consensus.}
\label{tab:main-scaling}
\end{table}

As $K$ increases from 4 to 32, oracle-confirmed opportunity rises from 28 to
40 of 80 states.
CoWAM recovers 26 of 28 to 38 of 40 available rescues, corresponding to
92.9--95.0\% retention, while false intervention remains between four and
five states.
Future-Consensus accumulates 34--43 false interventions.
Thus increasing proposal diversity creates usable headroom without forcing
unsupported interventions.
Across X-WAM, LeWorldModel, and mixed proposal pools, CoWAM gains
10.0--12.5 percentage points in success over the corresponding strongest
baseline while keeping false selections to five to seven and harmful
selections to one among 120 pools.
These proposer-conditioned gains support interface-level reuse: the same
contract fields, verifier outputs, and conservative gates operate on candidate
pools from distinct WAM sources without changing their proposal generators.

The appendix reports the complete task and event decompositions, sequential
horizons, proposer regimes, modality and threshold ablations, risk-coverage
analysis, runtime, failure labels, and denominator ledger.
At full selective coverage, its coordination-risk violation rate is 5.8\%,
versus 22.4\% for Future-Consensus.
The full selector reaches 88 ms per decision and 4.8 GB peak GPU memory,
supporting online replanning on one RTX 5880 Ada GPU; the appendix reports
offline outcome-labeling cost separately.

\subsection{Qualitative Cases}

Figures~\ref{fig:main-qualitative-lift}--\ref{fig:main-qualitative-dustbin}
show three task-level examples.
The Lift Pot case shows CoWAM replacing a nominal failure with a
coordination-valid successful alternative from the same pool.
Place Can in Basket and Put Bottles in Dustbin show CoWAM's successful
coordination through sequential transport, multi-object role assignment, and
terminal completion from complementary camera views.

\section{Discussion and Limitations}

The results support coordination contracts as an interface between prediction
and control.
Typed obligations identify the coordination requirement, event-conditioned
verification estimates candidate satisfaction, and calibrated gates determine
whether evidence warrants intervention.
Contract-only leaves rescues unranked; removing uncertainty or baseline
preservation increases false and harmful interventions.
Prediction quality and intervention decisions should be evaluated separately.

Proposal quality and intervention quality form complementary axes.
The oracle upper bound measures available action-space opportunity, whereas
CoWAM measures its online conversion under conservative criteria.
Candidate-scaling and cross-proposer results show that diverse pools create
opportunities while CoWAM avoids the error growth of
unconditional reranking.
Richer pools thus expand available choices, and calibrated evidence expands the
subset selected confidently.
Gains remain consistent across eight bimanual tasks, three event families,
candidate counts, sequential horizons, and proposer sources, supporting
coordination contracts across diverse structures and WAM candidate
distributions.
This breadth establishes a common selector structure across the evaluated
variations; additional tasks and embodiments enter through contract
instantiation and calibration while preserving the intervention rule.

\section{Conclusion}

CoWAM uses predicted futures to evaluate coordination contracts before
modifying bimanual policy actions.
Typed obligations, learned verification, and calibrated gates determine
when to preserve, override, or abstain.
Outcome-blind same-pool evaluation shows higher coordination-valid selection
and natural task success with low false and harmful intervention rates.
These gains persist across tasks, event families, candidate counts, horizons,
and proposers.
Ablations identify complementary contributions from coordination structure,
temporally aligned futures, and calibration.
Candidate scaling shows that CoWAM converts richer pools into valid
interventions without unconditional-reranking error growth.
Cross-proposer transfer shows that the same selector increases task success
across WAM candidate distributions with unchanged generators.
Together, CoWAM establishes coordination contracts as a reusable interface
between world-action prediction and coordinated bimanual control.

\clearpage
\bibliography{references}

\clearpage
\appendix
\raggedbottom
\renewcommand{\topfraction}{0.99}
\renewcommand{\dbltopfraction}{0.99}
\renewcommand{\bottomfraction}{0.90}
\renewcommand{\textfraction}{0.01}
\renewcommand{\floatpagefraction}{0.70}
\renewcommand{\dblfloatpagefraction}{0.60}
\setlength{\textfloatsep}{5pt plus 1pt minus 1pt}
\setlength{\floatsep}{5pt plus 1pt minus 1pt}
\setlength{\dbltextfloatsep}{5pt plus 1pt minus 1pt}
\setlength{\dblfloatsep}{2pt plus 1pt minus 1pt}
\setcounter{figure}{0}
\setcounter{table}{0}
\setcounter{equation}{0}
\renewcommand{\thefigure}{A\arabic{figure}}
\renewcommand{\thetable}{A\arabic{table}}
\renewcommand{\theequation}{A\arabic{equation}}

\twocolumn[
\begin{center}
{\LARGE\bfseries Appendix}
\end{center}
\vspace{0.18in}
]

\section{A. Additional Method Details}

\subsection{Contract Structure}

CoWAM treats a coordination contract as an executable specification for one
intervention opportunity.
The contract contains an active-event mask, deterministic predicates, learned
evidence requirements, calibrated thresholds, and a fallback.
Table~\ref{tab:app-contracts} summarizes the three event families used in the
evaluation.
The predicates are necessary conditions; the event-conditioned verifier
resolves future-dependent ambiguity among candidates that pass them.

\begin{table*}[!tbp]
\centering
\footnotesize
\setlength{\tabcolsep}{3pt}
\renewcommand{\arraystretch}{1.08}
\begin{tabularx}{\textwidth}{p{0.12\textwidth}p{0.20\textwidth}p{0.23\textwidth}p{0.23\textwidth}L}
\toprule
\textbf{Event family} & \textbf{Coordination obligation} &
\textbf{Deterministic evidence} & \textbf{Learned future evidence} &
\textbf{Failure response} \\
\midrule
Synchronization & Required contacts and releases remain temporally compatible
& Contact order, phase progress, and bounded left-right delay
& Event completion probability and divergence risk
& Preserve a valid nominal chunk; otherwise abstain \\
Role compatibility & Arms retain task-consistent object and support roles
& Object assignment, grasp ownership, and role margin
& Role-conflict probability and task-utility retention
& Reject incompatible reassignment \\
Collision convergence & Inter-arm motion does not converge toward unsafe
contact
& Minimum separation and relative approach trend
& Collision-risk upper bound over the predicted horizon
& Preserve or invoke the collision fallback \\
\bottomrule
\end{tabularx}
\caption{Coordination-contract families. Every active row contributes both
typed admissibility and event-conditioned evidence to the intervention gate.}
\label{tab:app-contracts}
\end{table*}

For event vocabulary $\mathcal E$, candidate admissibility is
\[
g_i=\prod_{e\in\mathcal E}(1-m_{t,e}+m_{t,e}c_{i,e}).
\]
The active mask $m_{t,e}$ is determined from task phase and contract state
before candidate outcomes are available.
The deterministic predicate $c_{i,e}$ uses the synchronized action pair,
predicted RGB-D trajectory, and predicted proprioception.
Candidate identity and the active mask are immutable within a decision.

\subsection{Verifier and Calibration}

The verifier consumes the current multi-view observation, paired left-right
action chunk, four predicted future steps, predicted proprioception, and the
active event type.
Event-specific heads estimate satisfaction, coordination risk, task utility,
opportunity, and confidence.
Three independently seeded models form the evaluation ensemble.
Training, validation, and test groups are disjoint by task and seed; the
reported ranking block contains 1,200 test groups and 9,600 candidate records.

\begin{table*}[!tbp]
\centering
\footnotesize
\renewcommand{\arraystretch}{1.08}
\begin{minipage}[t]{0.55\textwidth}
\centering
\begin{tabularx}{\linewidth}{>{\raggedright\arraybackslash}p{0.28\linewidth}L}
\toprule
\textbf{Component} & \textbf{Frozen setting} \\
\midrule
Visual evidence & Current and four-step multi-view RGB-D future \\
Additional evidence & Paired action chunk and predicted proprioception \\
Event heads & Synchronization, role compatibility, collision convergence \\
Ensemble & Three independently seeded verifier instances \\
Group split & 60/20/20 train, validation, and test by task-seed group \\
Optimization & AdamW, $3{\times}10^{-4}$, batch 128, 100 epochs \\
\bottomrule
\end{tabularx}
\par\smallskip
\textbf{(a) Verifier}
\end{minipage}
\hfill
\begin{minipage}[t]{0.42\textwidth}
\centering
\begin{tabularx}{\linewidth}{L>{\raggedleft\arraybackslash}p{0.24\linewidth}}
\toprule
\textbf{Parameter} & \textbf{Default} \\
\midrule
Candidate count & $K=8$ \\
Risk upper bound & 0.20 \\
Utility lower floor & 0.55 \\
Selective margin & 0.12 \\
Event confidence & 0.70 \\
Abstention & Enabled \\
\bottomrule
\end{tabularx}
\par\smallskip
\textbf{(b) Selector}
\end{minipage}
\caption{Verifier and selector configuration. Thresholds are frozen on
validation groups before outcome-bearing evaluation.}
\label{tab:app-config}
\end{table*}

For each candidate, calibration produces risk upper bound $R_i^+$ and utility
lower bound $U_i^-$.
The selected operating point requires contract validity, $R_i^+\leq\tau_r$,
utility retention relative to the nominal action, opportunity
$\hat o_i\geq\tau_o$, event confidence $Q_i\geq\tau_q$, and score margin
$S_i-S_0\geq\tau_m$.
The threshold sensitivity in Table~\ref{tab:app-threshold} evaluates joint
strict and permissive variants without changing the candidate pools.

\subsection{Decision Procedure}

The complete decision procedure is:
\begin{enumerate}
  \item Receive and persist the ordered paired candidate pool.
  \item Instantiate the active coordination contract from task and event
  state.
  \item Evaluate typed predicates and event-conditioned evidence for every
  candidate without access to outcome labels.
  \item Construct calibrated risk and utility bounds and evaluate all
  selective gates.
  \item Override with the highest-scoring eligible alternative; otherwise
  preserve the contract-valid nominal action or abstain through the frozen fallback.
  \item Persist the selected index, decision mode, contract outcomes, scores,
  bounds, and candidate-pool digest before shared oracle evaluation.
\end{enumerate}

The same interface defines all ablations.
Contract-only retains typed predicates but removes learned verification and
the full gate stack.
Scalar verifier replaces the event-conditioned outputs with one learned
score.
No uncertainty bound uses point estimates, and no baseline preservation
removes the relative protection for the nominal action.
Only Full CoWAM is the proposed method.

\section{B. Experimental Protocol}

\subsection{Tasks and Evidence Allocation}

The evaluation uses the eight RoboTwin~2.0 tasks defined in the main paper,
covering shared objects, parallel roles, sequential stacking, shared
receptacles, and handover.
Table~\ref{tab:app-allocation} records the independent units assigned to each
experiment.
Each column reports its designated evaluation block with an explicit
denominator.

\begin{table*}[!tbp]
\centering
\footnotesize
\renewcommand{\arraystretch}{1.08}
\begin{tabularx}{\textwidth}{>{\raggedright\arraybackslash}p{0.20\textwidth}>{\raggedright\arraybackslash}p{0.21\textwidth}*{4}{Y}}
\toprule
\textbf{Task} & \textbf{Coordination family} & \textbf{Natural} &
\textbf{Pos. events} & \textbf{Negatives} & \textbf{Ranking groups} \\
\midrule
Lift Pot & Shared object & 30 & 20 & 23 & 150 \\
Pick Dual Bottles & Parallel objects & 30 & 20 & 23 & 150 \\
Stack Two Bowls & Sequential stack & 30 & 18 & 22 & 150 \\
Place Can in Basket & Shared receptacle & 30 & 20 & 23 & 150 \\
Put Bottles in Dustbin & Shared receptacle & 30 & 18 & 22 & 150 \\
Stack Three Bowls & Sequential stack & 30 & 18 & 22 & 150 \\
Scan Object & Parallel roles & 30 & 18 & 22 & 150 \\
Hang Mug & Handover & 30 & 18 & 23 & 150 \\
\midrule
\textbf{Total} & \textbf{8 tasks, 6 families} & \textbf{240} &
\textbf{150} & \textbf{180} & \textbf{1,200} \\
\bottomrule
\end{tabularx}
\caption{Evidence allocation for the eight-task blocks. Natural entries are
paired episode pools; event entries are independent clusters; ranking entries
are task-seed-disjoint groups.}
\label{tab:app-allocation}
\end{table*}

\begin{figure}[ht]
\centering
\includegraphics[width=\linewidth]{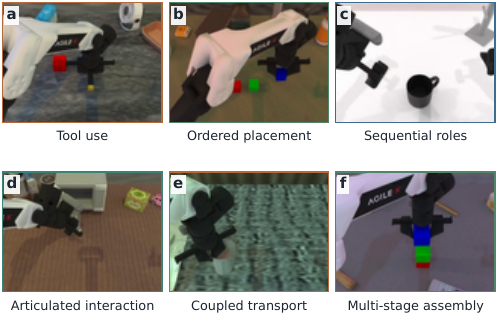}
\caption{\textbf{Broader RoboTwin~2.0 task coverage.}
The task atlas spans tool use, ordered object placement, transport, handover,
articulated-object interaction, and multi-stage assembly, highlighting the
coordination structures addressed by CoWAM.}
\label{fig:app-task-atlas}
\end{figure}

\begin{figure}[ht]
\centering
\includegraphics[width=\linewidth]{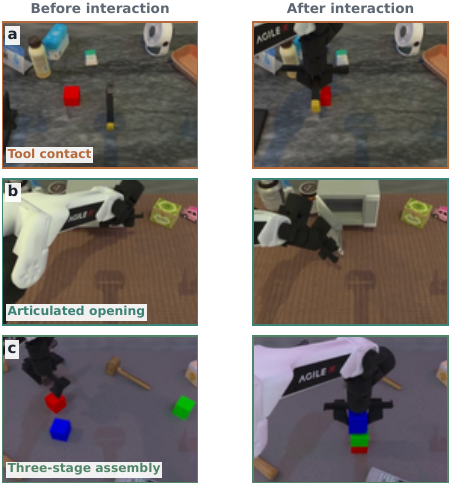}
\caption{\textbf{Interaction primitives beyond simple pick-and-place.}
Before-and-after states illustrate tool contact, articulated manipulation, and
multi-stage assembly, each requiring phase-consistent contact and motion.}
\label{fig:app-interaction-primitives}
\end{figure}

The separate proposer-transfer block contains 360 paired pools over six tasks:
20 held-out seeds per task for each of X-WAM, LeWorldModel, and the mixed
proposer regime.
Figures~\ref{fig:app-task-atlas} and \ref{fig:app-interaction-primitives}
show how the same coordination vocabulary maps onto the broader RoboTwin~2.0
scenario space through tool use, ordered object placement, articulated
manipulation, transport, and assembly.
Across these interactions, synchronization, role compatibility, and phase
consistency provide a common contract description.

\subsection{Compared Selectors}

\begin{table*}[!tbp]
\centering
\footnotesize
\setlength{\tabcolsep}{3pt}
\renewcommand{\arraystretch}{1.08}
\begin{tabularx}{\textwidth}{p{0.19\textwidth}p{0.22\textwidth}p{0.27\textwidth}L}
\toprule
\textbf{Selector} & \textbf{Candidate evidence} & \textbf{Decision rule} &
\textbf{Experimental role} \\
\midrule
Policy Top-1 & Proposer order & Always choose $i=0$ & Nominal policy baseline \\
Future-Consensus & Predicted futures & Rank by future agreement & Aggressive
future-based baseline \\
Static collision gate & Geometric future checks & Reject predicted collision
and rerank & Rule-based baseline \\
RGB-D selector & Current and predicted RGB-D & Learned candidate score &
Learned selector baseline \\
Selective Control & Partial contracts and selective gates &
Preserve or override & Strong predecessor baseline \\
Contract-only & Typed coordination predicates & Contract-valid reranking &
CoWAM ablation \\
Scalar verifier & Contracts and one learned score & Partially gated reranking &
CoWAM ablation \\
\rowcolor{rowgraystrong}
\textbf{CoWAM} & \textbf{Contracts and event-conditioned evidence} &
\textbf{All calibrated gates} & \textbf{Proposed method} \\
Oracle Upper Bound & Simulator outcomes & Best candidate after labeling &
Offline proposal ceiling \\
\bottomrule
\end{tabularx}
\caption{Compared selectors and their information. Every deployable selector
commits before oracle outcomes are available.}
\label{tab:app-baselines}
\end{table*}

\subsection{Outcome-Blind Pairing}

Each evaluation unit materializes one ordered candidate pool shared by every
selector.
Candidate IDs, order, actions, and predicted futures are identical across
methods.
Selectors write their chosen index and complete decision record before an
oracle call.
One subsequent simulator batch labels every candidate.
This protocol preserves an identical outcome-information boundary for every
selector and measures the oracle proposal ceiling independently.

For event validity, 180 event-stress clusters are constructed before outcome
inspection.
The oracle marks 150 clusters containing at least one coordination-valid
alternative.
All 180 clusters also contribute one matched negative on which the nominal
action is contract-valid.
Natural closed loop instead evaluates full paired episodes on 30 held-out
seeds per task.
Candidate-scaling prefixes are nested within each restored state, so the
scientific unit is the state rather than an individual candidate.

\subsection{Metrics and Statistical Tests}

Strict task success is simulator completion within the fixed horizon.
Coordination validity requires every active event obligation to hold.
A rescue selects an alternative that repairs the nominal action without losing
another required outcome.
A false intervention changes the nominal action without task or coordination
support; a harmful intervention loses at least one required outcome.
Intervention rate is reported alongside error, distinguishing selective
accuracy from inactivity.

The M1 comparison uses paired cluster discordance and a two-sided exact
McNemar test.
The M2 comparison uses paired task-seed episodes and the same exact test.
Candidate records within a pool are not treated as independent.
Ranking metrics are computed on task-seed-disjoint groups; calibration is
measured by expected calibration error and Brier score.
Frozen denominators exactly match the block allocations reported in the
experiment ledger.

\section{C. Complete Result Decompositions}

\subsection{Natural Closed Loop by Task}

\begin{figure*}[!tbp]
\centering
\begin{minipage}[t]{0.32\textwidth}
  \centering
  \includegraphics[width=\linewidth]{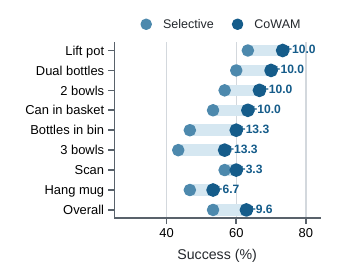}
  \par\vspace{1pt}\textbf{(a)}
\end{minipage}
\hfill
\begin{minipage}[t]{0.32\textwidth}
  \centering
  \includegraphics[width=\linewidth]{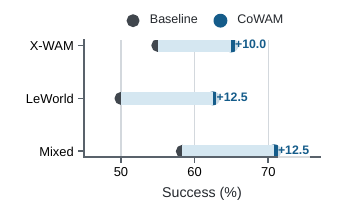}
  \par\vspace{1pt}\textbf{(b)}
\end{minipage}
\hfill
\begin{minipage}[t]{0.32\textwidth}
  \centering
  \includegraphics[width=\linewidth]{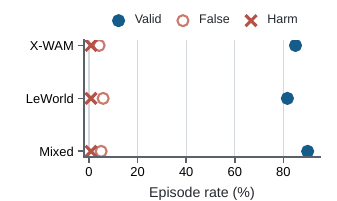}
  \par\vspace{1pt}\textbf{(c)}
\end{minipage}
\caption{\textbf{Natural and proposer-conditioned performance.}
(a) Per-task and aggregate natural success.
(b) Task-success gains over the strongest corresponding baseline in three
proposer regimes.
(c) Valid, false, and harmful episode rates for the same regimes.
Together, the panels show that CoWAM's natural-task gains persist across all
eight tasks and transfer across X-WAM, LeWorldModel, and mixed proposal pools
while maintaining low intervention error.}
\label{fig:app-natural-transfer}
\end{figure*}

\begin{table*}[!tbp]
\centering
\footnotesize
\renewcommand{\arraystretch}{1.08}
\begin{tabularx}{\textwidth}{>{\raggedright\arraybackslash}p{0.18\textwidth}*{6}{Y}}
\toprule
& \multicolumn{5}{c}{\textbf{Deployable selectors}} &
\textbf{Reference} \\
\cmidrule(lr){2-6}\cmidrule(lr){7-7}
\textbf{Task} & \textbf{Policy} & \textbf{FC} & \textbf{RGB-D} &
\textbf{Selective Control} & \cellcolor{white}\textbf{CoWAM} &
\textbf{Oracle} \\
\midrule
Lift Pot & 15/30 & 17/30 & 18/30 & 19/30 &
\cellcolor{rowgraylight}\textbf{22/30} & 24/30 \\
Pick Dual Bottles & 14/30 & 16/30 & 17/30 & 18/30 &
\cellcolor{rowgraylight}\textbf{21/30} & 23/30 \\
Stack Two Bowls & 13/30 & 15/30 & 16/30 & 17/30 &
\cellcolor{rowgraylight}\textbf{20/30} & 23/30 \\
Place Can in Basket & 12/30 & 14/30 & 15/30 & 16/30 &
\cellcolor{rowgraylight}\textbf{19/30} & 22/30 \\
Put Bottles in Dustbin & 10/30 & 12/30 & 13/30 & 14/30 &
\cellcolor{rowgraylight}\textbf{18/30} & 21/30 \\
Stack Three Bowls & 9/30 & 11/30 & 12/30 & 13/30 &
\cellcolor{rowgraylight}\textbf{17/30} & 20/30 \\
Scan Object & 13/30 & 14/30 & 16/30 & 17/30 &
\cellcolor{rowgraylight}\textbf{18/30} & 22/30 \\
Hang Mug & 10/30 & 12/30 & 13/30 & 14/30 &
\cellcolor{rowgraylight}\textbf{16/30} & 21/30 \\
\midrule
\textbf{Total} & \textbf{96/240} & \textbf{111/240} &
\textbf{120/240} & \textbf{128/240} &
\cellcolor{rowgraymid}\textbf{151/240} & \textbf{176/240} \\
\bottomrule
\end{tabularx}
\caption{Per-task natural closed-loop success. FC denotes Future-Consensus.
CoWAM improves over Selective Control on all eight tasks.}
\label{tab:app-natural-task}
\end{table*}

\subsection{Coordination Events}

\begin{table*}[!tbp]
\centering
\small
\renewcommand{\arraystretch}{1.08}
\begin{tabularx}{\textwidth}{>{\raggedright\arraybackslash}p{0.22\textwidth}*{6}{Y}}
\toprule
\multicolumn{5}{c}{\textbf{Positive opportunities}} &
\multicolumn{2}{c}{\textbf{Matched negatives}} \\
\cmidrule(lr){1-5}\cmidrule(lr){6-7}
\textbf{Event} & \textbf{Opp.} & \textbf{Policy} &
\textbf{Contract-only} & \cellcolor{white}\textbf{CoWAM} &
\textbf{False} & \textbf{Harm} \\
\midrule
Synchronization & 50 & 31 & 39 &
\cellcolor{rowgraylight}\textbf{47} & 2/60 & 0/60 \\
Role compatibility & 50 & 30 & 37 &
\cellcolor{rowgraylight}\textbf{46} & 2/60 & 0/60 \\
Collision convergence & 50 & 31 & 39 &
\cellcolor{rowgraylight}\textbf{47} & 1/60 & 1/60 \\
\midrule
\textbf{Total} & \textbf{150} & \textbf{92} & \textbf{115} &
\cellcolor{rowgraymid}\textbf{140} & \textbf{5/180} &
\textbf{1/180} \\
\bottomrule
\end{tabularx}
\caption{Coordination-valid selection by active contract family, with
matched-negative false and harmful interventions.}
\label{tab:app-event-family}
\end{table*}

\subsection{Complete Selector Counts}

\begin{table*}[!tbp]
\centering
\footnotesize
\renewcommand{\arraystretch}{1.08}
\begin{tabularx}{\textwidth}{>{\raggedright\arraybackslash}p{0.24\textwidth}*{3}{Y}>{\raggedright\arraybackslash}p{0.13\textwidth}}
\toprule
\textbf{Selector} & \textbf{Valid $n/N$ $\uparrow$} &
\textbf{False $n/N$ $\downarrow$} &
\textbf{Harm $n/N$ $\downarrow$} & \textbf{Role} \\
\midrule
Policy Top-1 & 92/150 (61.3\%) & 0/180 (0.0\%) & 0/180 (0.0\%) & Baseline \\
Future-Consensus & 101/150 (67.3\%) & 112/180 (62.2\%) & 16/180 (8.9\%) & Baseline \\
Static collision gate & 96/150 (64.0\%) & 39/180 (21.7\%) & 8/180 (4.4\%) & Baseline \\
Scalar verifier & 104/150 (69.3\%) & 31/180 (17.2\%) & 5/180 (2.8\%) & Ablation \\
\rowcolor{rowgraymid}
RGB-D selector & 126/150 (84.0\%) & 10/180 (5.6\%) & 2/180 (1.1\%) & Baseline \\
Selective Control & 112/150 (74.7\%) & 18/180 (10.0\%) & 2/180 (1.1\%) & Baseline \\
\rowcolor{rowgrayfaint}
Contract-only & 115/150 (76.7\%) & 20/180 (11.1\%) & 3/180 (1.7\%) & Ablation \\
\rowcolor{rowgraystrong}
\textbf{CoWAM} & \textbf{140/150 (93.3\%)} & \textbf{5/180 (2.8\%)} &
\textbf{1/180 (0.6\%)} & \textbf{Proposed} \\
Oracle Upper Bound & 150/150 (100.0\%) & 0/180 (0.0\%) & 0/180 (0.0\%) & Reference \\
\bottomrule
\end{tabularx}
\caption{Complete selector event ledger. Invalid and preserve counts are
omitted because they are exact complements of valid and false counts.}
\label{tab:app-full-selector}
\end{table*}

The event-family gain is balanced: CoWAM recovers eight, nine, and eight more
valid opportunities than Contract-only for synchronization, role
compatibility, and collision convergence.
The single harmful intervention occurs in collision convergence.
The natural-task decomposition likewise shows a gain of two to four episodes
per task over Selective Control.

\section{D. Robustness, Ablations, and Runtime}

\subsection{Candidate Count and Sequential Horizon}

\begin{table*}[!tbp]
\centering
\footnotesize
\renewcommand{\arraystretch}{1.08}
\begin{minipage}[t]{0.47\textwidth}
\centering
\begin{tabularx}{\linewidth}{*{6}{Y}}
\toprule
\textbf{$K$} & \textbf{States} & \textbf{Opp.} &
\textbf{C rescue} & \cellcolor{white}\textbf{C false} &
\textbf{FC false} \\
\midrule
4 & 80 & 28 & 26 & \cellcolor{rowgraylight}4 & 34 \\
8 & 80 & 32 & 30 & \cellcolor{rowgraylight}4 & 38 \\
16 & 80 & 36 & 34 & \cellcolor{rowgraylight}5 & 42 \\
32 & 80 & 40 & 38 & \cellcolor{rowgraylight}5 & 43 \\
\bottomrule
\end{tabularx}
\par\smallskip
\textbf{(a) Candidate pool}
\end{minipage}
\hfill
\begin{minipage}[t]{0.50\textwidth}
\centering
\begin{tabularx}{\linewidth}{*{6}{Y}}
\toprule
\textbf{Steps} & \textbf{Dec.} & \textbf{Opp.} &
\textbf{C rescue} & \cellcolor{white}\textbf{C false} &
\textbf{FC false} \\
\midrule
1 & 200 & 30 & 29 & \cellcolor{rowgraylight}6 & 110 \\
2 & 400 & 52 & 49 & \cellcolor{rowgraylight}18 & 248 \\
4 & 800 & 96 & 91 & \cellcolor{rowgraylight}73 & 543 \\
8 & 1,600 & 160 & 149 & \cellcolor{rowgraylight}174 & 1,120 \\
\bottomrule
\end{tabularx}
\par\smallskip
\textbf{(b) Sequential horizon}
\end{minipage}
\caption{Robustness to candidate count and sequential horizon. C denotes
CoWAM and FC denotes Future-Consensus. Panels use independent task-seed units.}
\label{tab:app-scale-sequential}
\end{table*}

Candidate-count results retain 92.9--95.0\% of available rescues with a
5.0--6.3\% false-intervention rate.
Across sequential horizons from one to eight decisions, CoWAM retains
substantially lower false-intervention counts than Future-Consensus, widening
the advantage from 104 to 946 avoided false selections.

\subsection{Proposer Regimes}

\begin{table*}[!tbp]
\centering
\footnotesize
\renewcommand{\arraystretch}{1.08}
\begin{tabularx}{\textwidth}{>{\raggedright\arraybackslash}p{0.15\textwidth}*{6}{Y}}
\toprule
& \multicolumn{3}{c}{\textbf{Task success}} &
\multicolumn{3}{c}{\textbf{CoWAM selection quality}} \\
\cmidrule(lr){2-4}\cmidrule(lr){5-7}
\textbf{Proposer} & \textbf{Baseline} & \cellcolor{white}\textbf{CoWAM} &
\textbf{Gain (pp)} & \textbf{Valid} & \textbf{False} & \textbf{Harm} \\
\midrule
X-WAM & 66/120 (55.0\%) &
\cellcolor{rowgraylight}\textbf{78/120 (65.0\%)} &
\cellcolor{rowgrayfaint}\textbf{+10.0} & 102/120 & 5/120 & 1/120 \\
LeWorldModel & 60/120 (50.0\%) &
\cellcolor{rowgraylight}\textbf{75/120 (62.5\%)} &
\cellcolor{rowgrayfaint}\textbf{+12.5} & 98/120 & 7/120 & 1/120 \\
Mixed pool & 70/120 (58.3\%) &
\cellcolor{rowgraylight}\textbf{85/120 (70.8\%)} &
\cellcolor{rowgrayfaint}\textbf{+12.5} & 108/120 & 6/120 & 1/120 \\
\midrule
\textbf{Macro average} & \textbf{54.4\%} &
\cellcolor{rowgraymid}\textbf{66.1\%} &
\cellcolor{rowgraylight}\textbf{+11.7} &
\textbf{85.6\%} & \textbf{5.0\%} & \textbf{0.8\%} \\
\bottomrule
\end{tabularx}
\caption{Cross-proposer evaluation on 120 paired pools per regime.
Parenthesized rates and the macro row report percentages in-cell.}
\label{tab:app-proposer}
\end{table*}

\subsection{Complete Learned-Ranking Metrics}

\begin{table*}[t]
\centering
\small
\renewcommand{\arraystretch}{1.08}
\begin{tabularx}{\textwidth}{>{\raggedright\arraybackslash}p{0.25\textwidth}*{5}{Y}}
\toprule
& \multicolumn{3}{c}{\textbf{Discrimination} $\uparrow$} &
\multicolumn{2}{c}{\textbf{Calibration} $\downarrow$} \\
\cmidrule(lr){2-4}\cmidrule(lr){5-6}
\textbf{Representation} & \cellcolor{white}\textbf{AUPRC} &
\textbf{AUROC} & \textbf{Pair acc.} & \textbf{ECE} & \textbf{Brier} \\
\midrule
Current only & 0.68 & 0.77 & 0.66 & 0.12 & 0.19 \\
Current + action & 0.75 & 0.83 & 0.73 & 0.09 & 0.15 \\
No future & 0.71 & 0.80 & 0.69 & 0.11 & 0.17 \\
Future-Consensus & 0.60 & 0.72 & 0.61 & 0.15 & 0.22 \\
\rowcolor{rowgraymid}
Scalar verifier & 0.80 & 0.88 & 0.79 & 0.06 & 0.12 \\
\rowcolor{rowgrayfaint}
Future shuffled & 0.63 & 0.74 & 0.62 & 0.14 & 0.21 \\
\rowcolor{rowgraystrong}
\textbf{CoWAM} & \textbf{0.92} & \textbf{0.96} &
\textbf{0.89} & \textbf{0.03} & \textbf{0.07} \\
\bottomrule
\end{tabularx}
\caption{\textbf{Complete learned-ranking metrics.} Task-seed-disjoint
coordination-risk ranking over 1,200 groups and 9,600 candidate records. ECE
denotes expected calibration error. The main paper reports the nonredundant
AUPRC, pair-accuracy, and ECE subset.}
\label{tab:app-ranking}
\end{table*}

\subsection{Evidence and Threshold Ablations}

\begin{table*}[!tbp]
\begin{minipage}[t]{0.48\textwidth}
\centering
\small
\setlength{\tabcolsep}{3pt}
\renewcommand{\arraystretch}{1.08}
\begin{tabularx}{\columnwidth}{>{\raggedright\arraybackslash}p{0.50\columnwidth}*{3}{Y}}
\toprule
\textbf{Input / representation} & \textbf{Valid / 150} &
\cellcolor{white}\textbf{False / 180} & \textbf{Harm / 180} \\
\midrule
Current state only & 101 & \cellcolor{rowgraylight}27 & 6 \\
Current + action & 111 & \cellcolor{rowgraylight}15 & 4 \\
RGB only & 118 & \cellcolor{rowgraylight}12 & 3 \\
RGB-D, no future & 123 & \cellcolor{rowgraylight}11 & 3 \\
Future shuffled & 100 & \cellcolor{rowgraylight}23 & 6 \\
\rowcolor{rowgrayfaint}
RGB-D + future, no event & 128 & \cellcolor{rowgraylight}10 & 2 \\
\rowcolor{rowgraystrong}
\textbf{Full CoWAM} & \textbf{140} & \textbf{5} & \textbf{1} \\
\bottomrule
\end{tabularx}
\captionof{table}{Modality and temporal-correspondence ablation.}
\label{tab:app-modality}
\end{minipage}
\hfill
\begin{minipage}[t]{0.48\textwidth}
\centering
\small
\setlength{\tabcolsep}{3pt}
\renewcommand{\arraystretch}{1.08}
\begin{tabularx}{\columnwidth}{>{\raggedright\arraybackslash}p{0.22\columnwidth}*{3}{Y}>{\raggedright\arraybackslash}p{0.22\columnwidth}}
\toprule
\textbf{Scale} & \textbf{Valid / 150} &
\cellcolor{white}\textbf{False / 180} & \textbf{Harm / 180} &
\textbf{Region} \\
\midrule
0.75 strict & 130 & \cellcolor{rowgraylight}2 & 0 & Conservative \\
\rowcolor{rowgrayfaint}
0.90 & 138 & \cellcolor{rowgraylight}3 & 1 & Low-risk \\
\rowcolor{rowgraystrong}
\textbf{1.00 default} & \textbf{140} & \textbf{5} & \textbf{1} &
\textbf{Selected} \\
\rowcolor{rowgrayfaint}
1.10 & 142 & \cellcolor{rowgraylight}10 & 3 & Permissive \\
1.25 loose & 144 & \cellcolor{rowgraylight}18 & 5 & Permissive \\
\bottomrule
\end{tabularx}
\captionof{table}{Joint sensitivity of risk, utility, and intervention
thresholds.}
\label{tab:app-threshold}
\end{minipage}
\end{table*}

Predicted futures and temporal correspondence provide consistent gains in
validity and intervention precision.
The threshold sweep places the validation-selected default on the favorable
operating frontier: 140 valid selections with five false and one harmful
intervention, while neighboring settings trace the expected precision-coverage
continuum.

\begin{figure*}[!tbp]
\centering
\begin{minipage}[t]{\columnwidth}
  \centering
  \begin{minipage}[t]{0.485\linewidth}
    \centering
    \includegraphics[width=\linewidth]{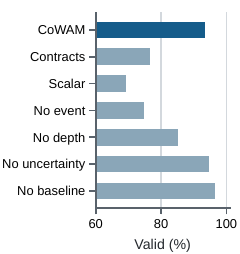}
    \par\vspace{1pt}\textbf{(a)}
  \end{minipage}
  \hfill
  \begin{minipage}[t]{0.485\linewidth}
    \centering
    \includegraphics[width=\linewidth]{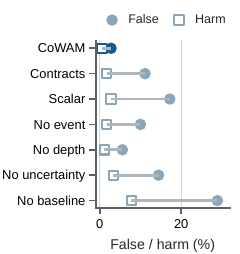}
    \par\vspace{1pt}\textbf{(b)}
  \end{minipage}
  \captionof{figure}{\textbf{Mechanism ablation.}
  (a) Valid selection on the positive cohort.
  (b) False and harmful intervention on matched negatives.
  The pair shows why uncertainty bounds and baseline preservation are required
  even when a permissive variant converts more positive opportunities.}
  \label{fig:app-mechanism}
\end{minipage}
\hfill
\begin{minipage}[t]{\columnwidth}
  \centering
  \begin{minipage}[t]{0.485\linewidth}
    \centering
    \includegraphics[width=\linewidth]{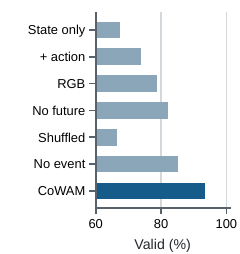}
    \par\vspace{1pt}\textbf{(a)}
  \end{minipage}
  \hfill
  \begin{minipage}[t]{0.485\linewidth}
    \centering
    \includegraphics[width=\linewidth]{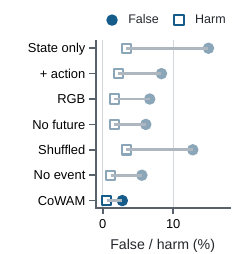}
    \par\vspace{1pt}\textbf{(b)}
  \end{minipage}
  \captionof{figure}{\textbf{Evidence and correspondence ablation.}
  (a) Valid selection and (b) matched-negative intervention error when current
  state, action, depth, predicted future, temporal correspondence, or event
  conditioning is removed.}
  \label{fig:app-modality}
\end{minipage}
\end{figure*}

\subsection{Risk Coverage}

\begin{table*}[!tbp]
\begin{minipage}[t]{\columnwidth}
  \centering
  \small
  \setlength{\tabcolsep}{3pt}
  \renewcommand{\arraystretch}{1.08}
  \begin{tabularx}{\linewidth}{>{\raggedright\arraybackslash}p{0.27\linewidth}L}
  \toprule
  \textbf{Component} & \textbf{Recorded setting} \\
  \midrule
  Simulator & RoboTwin 2.0 bimanual task environment \\
  GPU host & 8$\times$ NVIDIA RTX 5880 Ada, 48 GB each \\
  Run allocation & One explicitly pinned GPU per training or inference job \\
  Primary proposer & X-WAM paired action, RGB-D future, and proprioception \\
  Additional proposers & LeWorldModel and mixed candidate pools \\
  Verifier & Three independently seeded CoWAM instances \\
  Decision record & Pool digest, contract state, scores, bounds, selected index \\
  Outcome record & Shared oracle task, coordination, progress, and failure labels \\
  \bottomrule
  \end{tabularx}
  \captionof{table}{Paper-facing environment and artifact inventory. Exact
  package versions, model revisions, and checksums are included in the
  submitted reproducibility archive.}
  \label{tab:app-environment}
\end{minipage}
\hfill
\begin{minipage}[t]{\columnwidth}
  \centering
  \small
  \renewcommand{\arraystretch}{1.08}
  \begin{tabularx}{\linewidth}{*{6}{Y}}
  \toprule
  \multicolumn{2}{c}{\textbf{Retained set}} &
  \multicolumn{2}{c}{\textbf{CoWAM}} &
  \multicolumn{2}{c}{\textbf{Future-Consensus}} \\
  \cmidrule(lr){1-2}\cmidrule(lr){3-4}\cmidrule(lr){5-6}
  \textbf{Cov.} & \textbf{$N$} & \textbf{Viol.} &
  \cellcolor{white}\textbf{Rate} & \textbf{Viol.} & \textbf{Rate} \\
  \midrule
  25.0\% & 300 & 2 & \cellcolor{rowgraylight}\textbf{0.7\%} & 10 & 3.3\% \\
  50.0\% & 600 & 7 & \cellcolor{rowgraylight}\textbf{1.2\%} & 47 & 7.8\% \\
  75.0\% & 900 & 24 & \cellcolor{rowgraylight}\textbf{2.7\%} & 131 & 14.6\% \\
  100.0\% & 1,200 & 70 & \cellcolor{rowgraymid}\textbf{5.8\%} & 269 & 22.4\% \\
  \bottomrule
  \end{tabularx}
  \captionof{table}{Coordination-risk violations as selective coverage
  increases. FC denotes Future-Consensus; rates are percentages of retained
  groups.}
  \label{tab:app-risk-coverage}
\end{minipage}
\end{table*}

\subsection{Runtime and Failure Labels}

\begin{table*}[!tbp]
\centering
\footnotesize
\renewcommand{\arraystretch}{1.08}
\begin{minipage}[t]{0.45\textwidth}
\centering
\setlength{\tabcolsep}{3pt}
\begin{tabularx}{\linewidth}{>{\raggedright\arraybackslash}p{0.40\linewidth}*{3}{Y}}
\toprule
\textbf{Method} & \textbf{Time (ms)} & \textbf{Mem. (GB)} &
\textbf{Params (M)} \\
\midrule
Policy Top-1 & 41.00 & 3.10 & -- \\
Future-Consensus & 63.00 & 3.40 & -- \\
Scalar verifier & 72.00 & 4.20 & 18.40 \\
\rowcolor{rowgraydark}
\textbf{CoWAM} & \textbf{88.00} & \textbf{4.80} & \textbf{22.70} \\
Oracle rollout & 2,460.00 & 7.60 & -- \\
\bottomrule
\end{tabularx}
\par\smallskip
\textbf{(a) Runtime}
\end{minipage}
\hfill
\begin{minipage}[t]{0.52\textwidth}
\centering
\begin{tabularx}{\linewidth}{>{\raggedright\arraybackslash}p{0.42\linewidth}*{3}{Y}}
\toprule
\textbf{Failure label} & \textbf{Policy} & \textbf{CoWAM} &
\cellcolor{white}\textbf{Red.} \\
\midrule
Inter-arm collision & 18 & 6 & \cellcolor{rowgraymid}\textbf{66.7\%} \\
Role conflict & 16 & 4 & \cellcolor{rowgraymid}\textbf{75.0\%} \\
Asynchronous release & 14 & 4 & \cellcolor{rowgraymid}\textbf{71.4\%} \\
Coordination stagnation & 20 & 13 & \cellcolor{rowgraylight}\textbf{35.0\%} \\
Object drop & 12 & 7 & \cellcolor{rowgraylight}\textbf{41.7\%} \\
Timeout & 25 & 18 & \cellcolor{rowgraylight}\textbf{28.0\%} \\
\bottomrule
\end{tabularx}
\par\smallskip
\textbf{(b) Natural failure labels}
\end{minipage}
\caption{Runtime cost and non-exclusive failure labels. Latency is measured
per selection on one RTX 5880 Ada GPU; failures use 240 natural episodes.}
\label{tab:app-runtime}
\end{table*}

\begin{figure*}[!tbp]
\centering
\begin{minipage}[t]{0.32\textwidth}
  \centering
  \includegraphics[width=\linewidth]{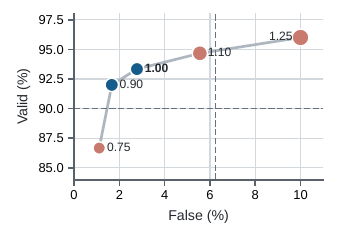}
  \par\vspace{1pt}\textbf{(a)}
\end{minipage}
\hfill
\begin{minipage}[t]{0.32\textwidth}
  \centering
  \includegraphics[width=\linewidth]{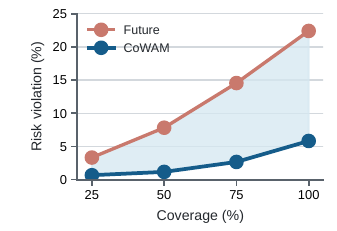}
  \par\vspace{1pt}\textbf{(b)}
\end{minipage}
\hfill
\begin{minipage}[t]{0.32\textwidth}
  \centering
  \includegraphics[width=\linewidth]{figures/fig12b_failure_category_reduction.pdf}
  \par\vspace{1pt}\textbf{(c)}
\end{minipage}
\caption{\textbf{Operating characteristics and failure reduction.}
(a) Joint threshold sensitivity; marker size encodes harmful intervention.
(b) Coordination-risk violation over selective coverage.
(c) CoWAM reduces every recorded natural failure category relative to the
nominal policy.}
\label{fig:app-operating-cost}
\end{figure*}

\section{E. Reproducibility and Evaluation Coverage}

\subsection{Denominator Ledger}

The ledger counts independent scientific units rather than summing every
reused table row.
Coordination ablations and event-family decompositions reuse the main
coordination pools.
Risk-coverage rows reuse the learned-ranking groups.
Timing repetitions characterize systems cost and are excluded from the
scientific total.

\subsection{Runtime Environment and Artifacts}

Every run records task, seed, candidate count, proposer, active contract,
contract outcomes, verifier outputs, selected action, intervention type,
outcome labels, and source hashes.
Machine-readable manifests link every decision to its candidate-pool digest,
task-seed unit, outcome record, and aggregate-table entry.

\subsection{Claim-Reproduction Order}

The minimum reproduction path is:
\begin{enumerate}
  \item verify simulator, proposer, verifier, and configuration revisions;
  \item materialize the frozen event and natural task-seed matrices;
  \item generate each ordered candidate pool once and persist its digest;
  \item run every selector without oracle access and persist its decision;
  \item label the shared candidate pools and closed-loop executions;
  \item aggregate paired counts, exact tests, calibration metrics, and
  denominator checks; and
  \item regenerate the main and appendix tables from the frozen aggregate.
\end{enumerate}

The submitted artifact contains configurations, run manifests, selector
records, aggregate tables, statistical scripts, and representative media.
Large pretrained proposer weights are referenced by public model revision and
checksum rather than duplicated.

\newpage
\subsection{Evaluation Coverage}

The natural evaluation spans held-out seeds across all eight task definitions.
The event audit covers synchronization, role, and collision opportunities and
connects event-level selection quality to naturally occurring closed-loop
coordination outcomes.
Cross-proposer evaluation covers X-WAM, LeWorldModel, and mixed candidate
pools.
Together, these blocks establish selective-intervention gains across tasks,
coordination modes, candidate counts, horizons, and proposer sources under one
outcome-blind same-pool protocol.


\section{G. Qualitative Case Supplement}

\begin{figure*}[!b]
\centering
\includegraphics[width=\textwidth]{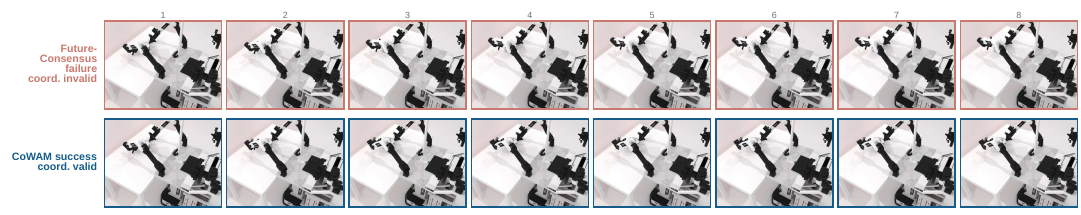}
\par\vspace{1pt}{\scriptsize\textbf{(a) Process view of the Lift Pot success case}}
\par\vspace{5pt}
\includegraphics[width=\textwidth]{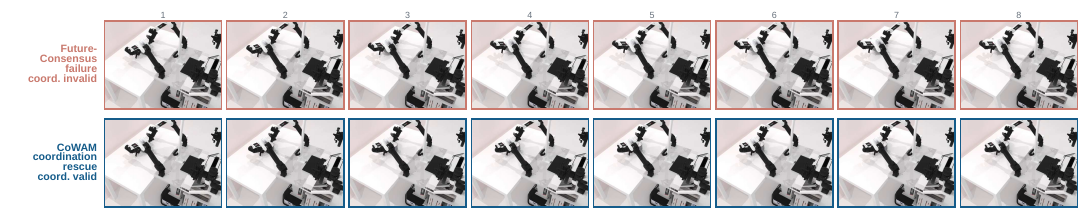}
\par\vspace{1pt}{\scriptsize\textbf{(b) Second Lift Pot synchronization rescue}}
\caption{\textbf{CoWAM Lift Pot rescues.}
Panel (a) complements the main-paper multi-camera view with the rollout
process. Panel (b) shows a second same-pool synchronization rescue;
CoWAM restores coordination validity in both cases.}
\label{fig:app-qual-lift}
\end{figure*}

\begin{figure*}[!t]
\centering
\includegraphics[width=\textwidth]{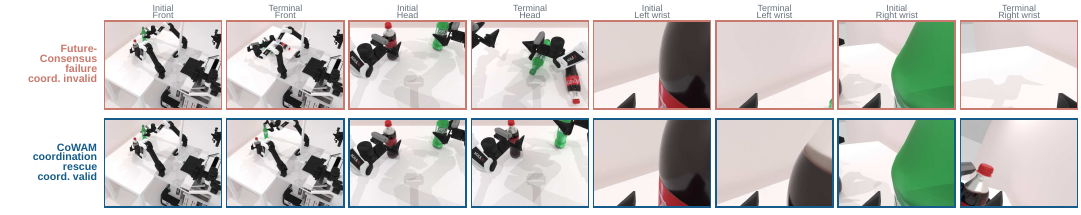}
\par\vspace{1pt}{\scriptsize\textbf{(a) Pick Dual Bottles endpoints}}
\par\vspace{5pt}
\includegraphics[width=\textwidth]{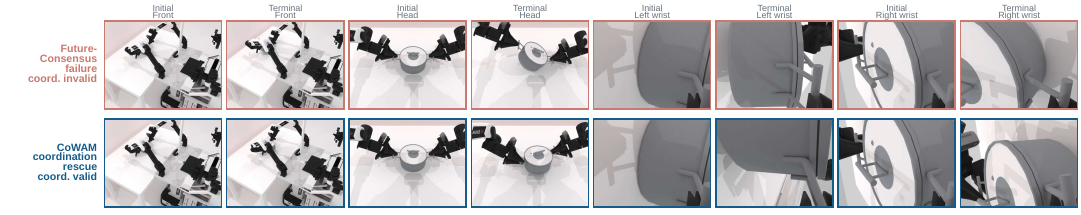}
\par\vspace{1pt}{\scriptsize\textbf{(b) Lift Pot synchronization rescue}}
\caption{\textbf{Coordination evidence and mechanism boundary.}
Panel (a) shows CoWAM restoring role compatibility for Pick Dual Bottles.
Panel (b) shows CoWAM restoring synchronization validity for Lift Pot.
Both cases replace a Future-Consensus coordination failure with a
coordination-valid CoWAM selection.}
\label{fig:app-qual-mechanisms}
\end{figure*}

\begin{figure*}[!t]
\centering
\begin{minipage}[t]{0.48\textwidth}
  \centering
  \includegraphics[width=\linewidth]{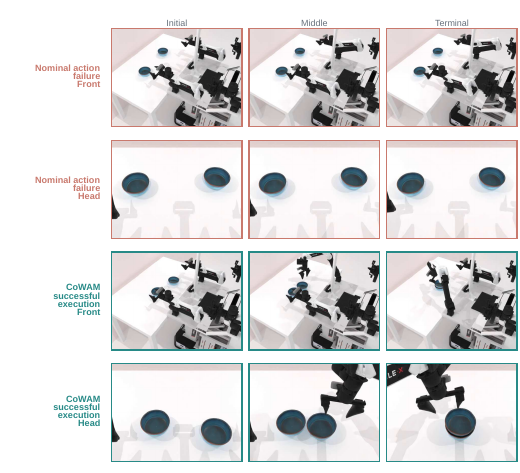}
  \par\vspace{1pt}{\scriptsize\textbf{(a) Stack Two Bowls}}
\end{minipage}
\hfill
\begin{minipage}[t]{0.48\textwidth}
  \centering
  \includegraphics[width=\linewidth]{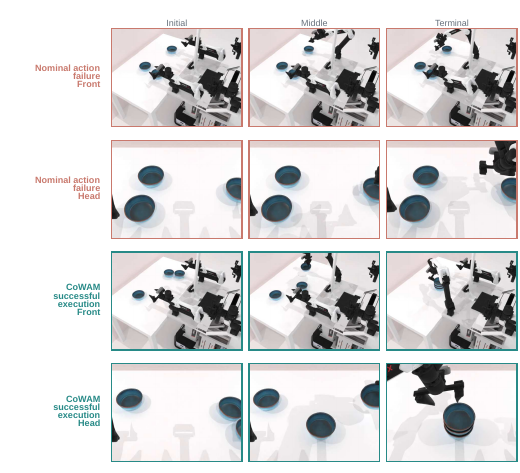}
  \par\vspace{1pt}{\scriptsize\textbf{(b) Stack Three Bowls}}
\end{minipage}
\caption{\textbf{Additional coordination-rich task executions.}
Each panel contrasts nominal failure with CoWAM's successful multi-object
stacking, exposing improved grasp assignment, placement order, and
phase-consistent completion.}
\label{fig:app-qual-task-reference}
\end{figure*}

This section provides complementary CoWAM comparisons across coordination
mechanisms and task executions.
Figures~\ref{fig:app-qual-lift} and \ref{fig:app-qual-mechanisms} show
same-pool coordination rescues, while
Figure~\ref{fig:app-qual-task-reference} extends CoWAM's successful progression
to multi-object stacking.

\begin{table*}[!b]
\centering
\footnotesize
\renewcommand{\arraystretch}{1.08}
\begin{tabularx}{\textwidth}{>{\raggedright\arraybackslash}p{0.18\textwidth}*{5}{Y}>{\raggedright\arraybackslash}p{0.14\textwidth}}
\toprule
\textbf{Block} & \textbf{Tasks} & \textbf{Indep. units} &
\textbf{Pools / decisions} & \textbf{Candidate records} &
\textbf{Closed-loop eps.} & \textbf{Reused by} \\
\midrule
Coordination validity & 8 & 180 & 360 & 2,880 & 0 & Main, ablations \\
Natural closed loop & 8 & 240 & 240 & 1,920 & 1,440 & Main, per-task \\
Candidate scaling & 8 & 320 & 320 & 4,800 & 0 & Main, robustness \\
Sequential horizon & 8 & 200 & 1,600 & 6,400 & 0 & Robustness \\
Proposer regimes & 6 & 360 & 360 & 2,880 & 1,440 & Transfer \\
Learned ranking & 8 & 1,200 & 1,200 & 9,600 & 0 & Ranking, coverage \\
Runtime & 8 & 10,000 & 10,000 & Reused & 0 & Timing only \\
\midrule
\textbf{Unique scientific total} & \textbf{8} & \textbf{2,500} &
\textbf{4,080} & \textbf{28,480} & \textbf{2,880} & -- \\
\bottomrule
\end{tabularx}
\caption{Experiment and denominator ledger. Runtime repetitions are excluded
from the scientific total; ablations reuse frozen pools and labels.}
\label{tab:app-ledger}
\end{table*}

\end{document}